\documentclass{article}

\usepackage{times}
\usepackage{graphicx} 
\usepackage{caption,subcaption}

\usepackage{natbib}

\usepackage{algorithm}
\usepackage{algorithmic}

\usepackage{booktabs}
\usepackage{multirow}

\usepackage{url}

\usepackage{amsmath,amssymb,dsfont}

\usepackage{hyperref}

\usepackage[accepted]{icml2016}

\icmltitlerunning{Bayesian Poisson Tucker Decomposition for Learning
  the Structure of International Relations}

\newcommand{\ija}{i {\xrightarrow{a}} j}
\newcommand{\yijat}{y^{(t)}_{\ija}}
\newcommand{\cdk}{c {\xrightarrow{k}} d}

\newcommand{\btheta}{\boldsymbol{\theta}}
\newcommand{\bphi}{\boldsymbol{\phi}}

\newcommand{\g}{\,|\,}
\newcommand{\teq}{\!=\!}

\newcommand{\Pois}{\textrm{Po}}

\newcommand{\rotcircle}{\,\mathbin{\text{\rotatebox[origin=c]{90}{$\circlearrowright$}}}}

\begin{document}

\twocolumn[
  \icmltitle{Bayesian Poisson Tucker Decomposition\\
    for Learning the Structure of International Relations}

\icmlauthor{Aaron Schein}{aschein@cs.umass.edu}
\icmladdress{University of Massachusetts Amherst}
\icmlauthor{Mingyuan Zhou}{mingyuan.zhou@mccombs.utexas.edu}
\icmladdress{University of Texas at Austin}
\icmlauthor{David M. Blei}{david.blei@columbia.edu}
\icmladdress{Columbia University}
\icmlauthor{Hanna Wallach}{wallach@microsoft.com}
\icmladdress{Microsoft Research New York City}

\icmlkeywords{international relations, computational social science,
  networks, community detection, matrix factorization, tensor
  decomposition, Tucker decomposition, Bayesian nonparametrics, topic
  modeling, MCMC, latent Dirichlet allocation}

\vskip 0.3in
]

\begin{abstract}
  We introduce Bayesian Poisson Tucker decomposition (BPTD) for
  modeling country--country interaction event data. These data consist
  of interaction events of the form ``country $i$ took action $a$
  toward country $j$ at time $t$.'' BPTD discovers overlapping
  country--community memberships, including the number of latent
  communities. In addition, it discovers directed community--community
  interaction networks that are specific to ``topics'' of action types
  and temporal ``regimes.'' We show that BPTD yields an efficient MCMC
  inference algorithm and achieves better predictive performance than
  related models. We also demonstrate that it discovers interpretable
  latent structure that agrees with our knowledge of international
  relations.
\end{abstract}

\section{Introduction}
\label{sec:introduction}

Like their inhabitants, countries interact with one another: they
consult, negotiate, trade, threaten, and fight. These interactions are
seldom uncoordinated. Rather, they are connected by a fabric of
overlapping communities, such as security coalitions, treaties, trade
cartels, and military alliances. For example, OPEC coordinates the
petroleum export policies of its thirteen member countries, LAIA
fosters trade among Latin American countries, and NATO guarantees
collective defense against attacks by external parties.

A single country can belong to multiple communities, reflecting its
different identities. For example, Venezuela---an oil-producing
country and a Latin American country---is a member of both OPEC and
LAIA. When Venezuela interacts with other countries, it sometimes does
so as an OPEC member and sometimes does so as a LAIA member.

Countries engage in both within-community and between-community
interactions. For example, when acting as an OPEC member, Venezuela
consults with other OPEC countries, but trades with non-OPEC,
oil-importing countries. Moreover, although Venezuela engages in
between-community interactions when trading as an OPEC member, it
engages in within-community interactions when trading as a LAIA
member. To understand or predict how countries interact, we must
account for their community memberships and how those memberships
influence their actions.

In this paper, we take a new approach to learning overlapping
communities from interaction events of the form ``country $i$ took
action $a$ toward country $j$ at time $t$.'' A data set of such
interaction events can be represented as either 1) a set of event
tokens, 2) a tensor of event type counts, or 3) a series of weighted
multinetworks. Models that use the token representation naturally
yield efficient inference algorithms, models that use the tensor
representation exhibit good predictive performance, and models that
use the network representation learn latent structure that aligns with
well-known concepts such as communities. Previous models of
interaction event data have each used a subset of these
representations. Our approach---Bayesian Poisson Tucker decomposition
(BPTD)---takes advantage of all three.

\pagebreak

\begin{figure}[t]
\centering
\vspace{-1.7em}
\begin{subfigure}[t]{0.54\columnwidth}
\caption*{}
\includegraphics[width=\columnwidth]{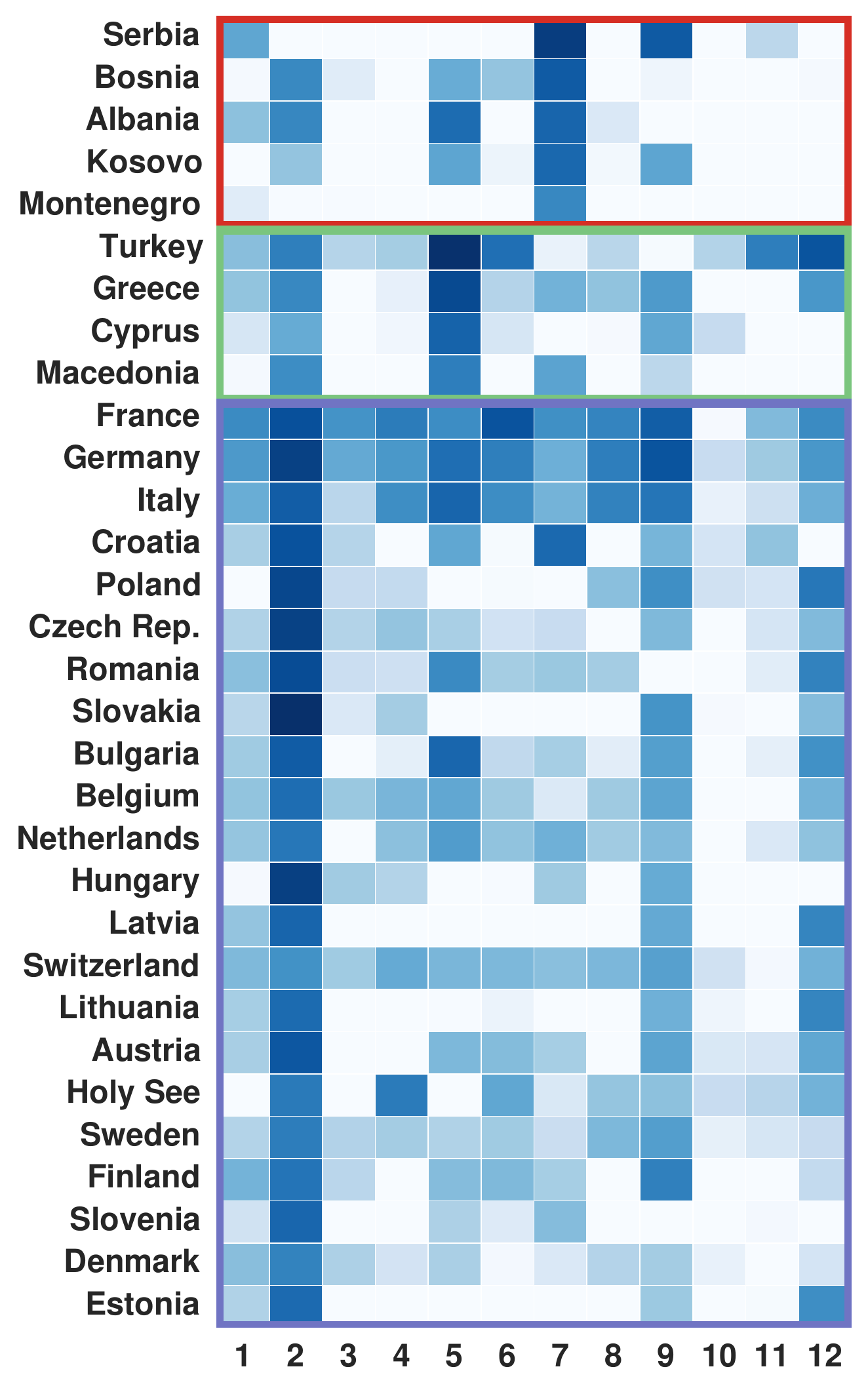}
\end{subfigure}
\begin{subfigure}[t]{0.44\columnwidth}
\caption*{}
\includegraphics[width=\columnwidth]{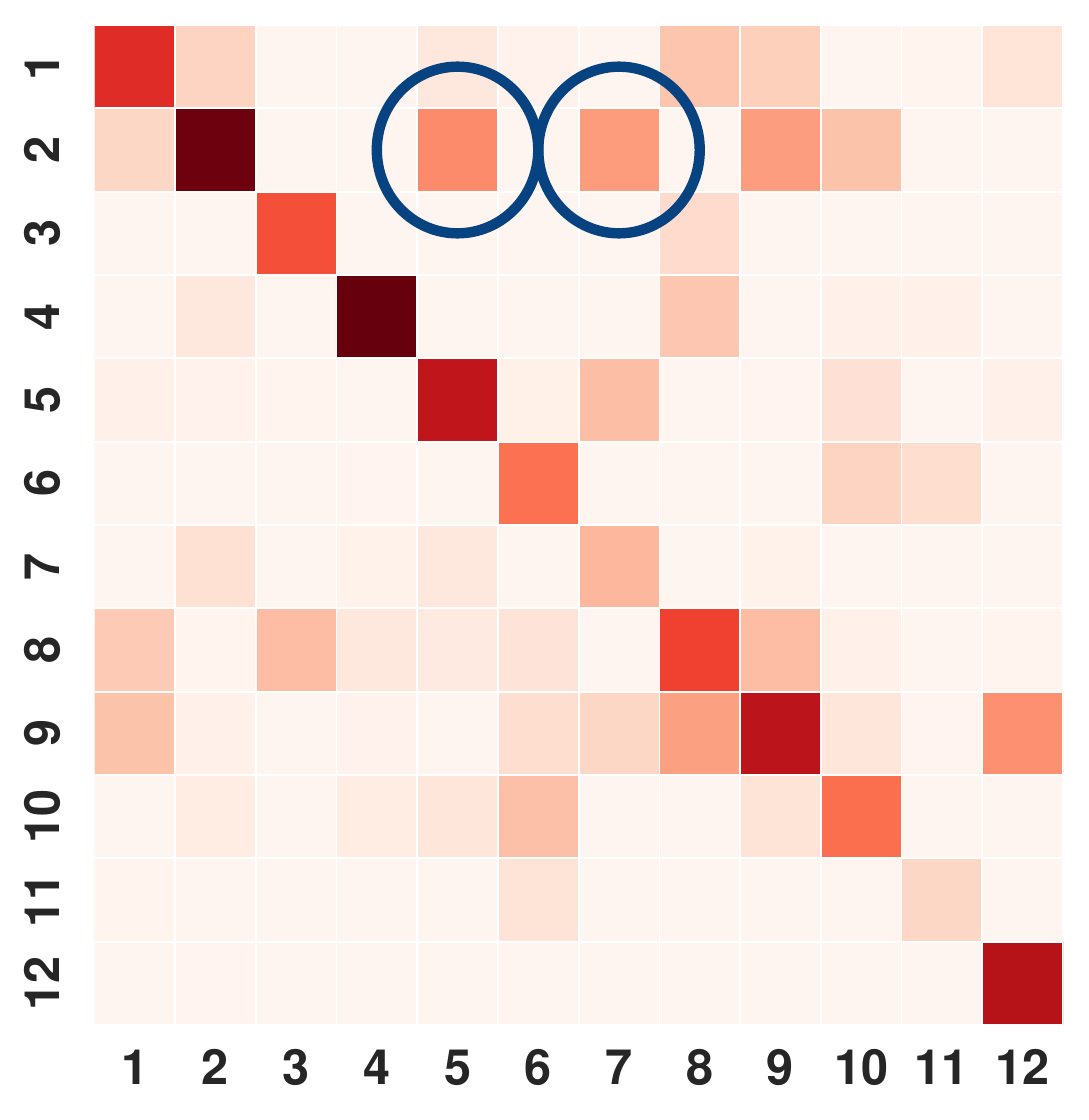}
\includegraphics[width=\columnwidth]{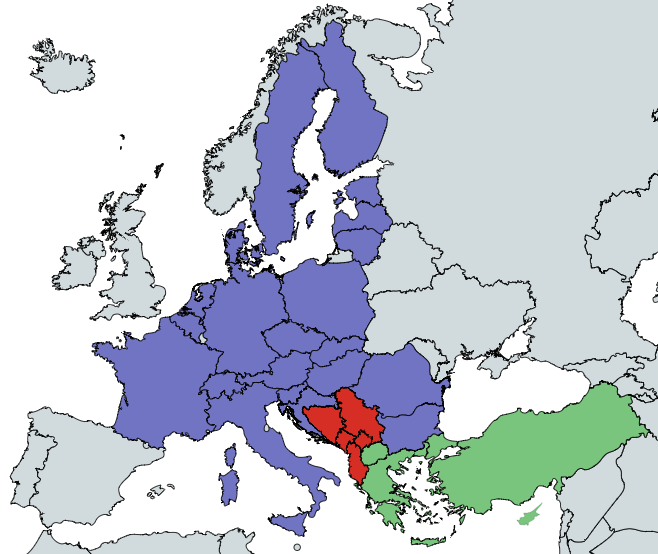}
\end{subfigure}
\vspace{0.3em}
\caption{Latent structure learned by BPTD from country--country
  interaction events between 1995 and 2000. \emph{Top right}: A
  community--community interaction network specific to a single topic
  of action types and temporal regime. The topic places most of its
  mass on the \emph{Intend to Cooperate} and \emph{Consult} actions,
  so this network represents cooperative community--community
  interactions. The two strongest between-community interactions
  (circled) are 2${\xrightarrow{}}$5 and
  2${\xrightarrow{}}$7. \emph{Left}: Each row depicts the overlapping
  community memberships for a single country. We show only those
  countries whose strongest community membership is to either
  community 2, 5, or 7. We ordered the countries
  accordingly. Countries strongly associated with community 7 are at
  highlighted in red; countries associated with community 5 are
  highlighted in green; and countries associated with community 2 are
  highlighted in purple. \emph{Bottom right}: Each country is colored
  according to its strongest community membership. The latent
  communities have a very strong geographic interpretation.}
\label{fig:example}
\vskip -5pt
\end{figure}

BPTD builds on the classic Tucker
decomposition~\citep{tucker64extension} to factorize a tensor of event
type counts into three factor matrices and a four-dimensional core
tensor (section~\ref{sec:model}). The factor matrices embed countries
into communities, action types into ``topics,'' and time steps into
``regimes.'' The core tensor interacts communities, topics, and
regimes. The country--community factors enable BPTD to learn
overlapping community memberships, while the core tensor enables it to
learn directed community--community interaction networks specific to
topics of action types and temporal regimes. Figure~\ref{fig:example}
illustrates this structure. BPTD leads to an efficient MCMC inference
algorithm (section~\ref{sec:inference}) and achieves better predictive
performance than related models
(section~\ref{sec:prediction}). Finally, BPTD discovers interpretable
latent structure that agrees with our knowledge of international
relations (section~\ref{sec:exploration}).

\section{Bayesian Poisson Tucker Decomposition}
\label{sec:model}

We can represent a data set of interaction events as a set of $N$
event tokens, where a single token $e_n = (\ija, t)$ indicates that
sender country $i \in [V]$ took action $a \in [A]$ toward receiver
country $j \in [V]$ during time step $t \in [T]$. Alternatively, we
can aggregate these event tokens into a four-dimensional tensor
$\boldsymbol{Y}$, where element $\yijat$ is a count of the number of
events of type $(\ija, t)$. This tensor will be sparse because most
event types never actually occur in practice. Finally, we can
equivalently view this count tensor as a series of $T$ weighted
multinetwork snapshots, where the weight on edge $\ija$ in the
$t^{\textrm{th}}$ snapshot is $\yijat$.

BPTD models each element of count tensor $\boldsymbol{Y}$ as
\begin{equation}
  \yijat \sim \Pois\!\left( \sum_{c=1}^C \theta_{ic} \!\sum_{d=1}^C
  \theta_{jd} \!\sum_{k=1}^K \phi_{ak} \!\sum_{r=1}^R
  \psi_{tr}\,\lambda^{(r)}_{\cdk}\right),
  \label{eqn:likelihood}
\end{equation}
where $\theta_{ic}$, $\theta_{jd}$, $\phi_{ak}$, $\psi_{tr}$, and
$\lambda^{(r)}_{\cdk}$ are positive real numbers. Factors
$\theta_{ic}$ and $\theta_{jd}$ capture the rates at which countries
$i$ and $j$ participate in communities $c$ and $d$, respectively;
factor $\phi_{ak}$ captures the strength of association between action
$a$ and topic $k$; and $\psi_{tr}$ captures how well regime $r$
explains the events in time step $t$. We can collectively view the $V
\times C$ country--community factors as a latent factor matrix
$\Theta$, where the $i^{\textrm{th}}$ row represents country $i$'s
community memberships. Similarly, we can view the $A \times K$
action--topic factors and the $T \times R$ time-step--regime factors
as latent factor matrices $\Phi$ and $\Psi$, respectively. Factor
$\lambda_{\cdk}^{(r)}$ captures the rate at which community $c$ takes
actions associated with topic $k$ toward community $d$ during regime
$r$. The $C \times C \times K \times R$ such factors form a core
tensor $\boldsymbol{\Lambda}$ that interacts communities, topics, and
regimes.

The country--community factors are gamma-distributed,
\vspace{-0.4em}
\begin{equation}
  \theta_{ic} \sim \Gamma\!\left(\alpha_i, \beta_i \right),
\vspace{-0.4em}
\end{equation}
where the shape and rate parameters $\alpha_i$ and $\beta_i$ are specific
to country $i$. We place an uninformative gamma prior over these shape
and rate parameters: $\alpha_i, \beta_i \sim \Gamma\!\left(\epsilon_0,
\epsilon_0\right)$. This hierarchical prior enables BPTD to express
heterogeneity in the countries' rates of activity. For example, we
expect that the US will engage in more interactions than Burundi.

The action--topic and time-step--regime factors are also
gamma-distributed; however, we assume that these factors are drawn
directly from an uninformative gamma prior,
\vspace{-0.4em}
\begin{equation}
  \phi_{ak}, \psi_{tr} \sim \Gamma\!\left(\epsilon_0, \epsilon_0
  \right).
  \vspace{-0.4em}
\end{equation}

Because BPTD learns a single embedding of countries into communities,
it preserves the traditional network-based notion of community
membership. Any sender--receiver asymmetry is captured by the core
tensor $\boldsymbol{\Lambda}$, which we can view as a compression of
count tensor $\boldsymbol{Y}$. By allowing on-diagonal elements, which
we denote by $\lambda_{c\rotcircle^k}^{(r)}$ and off-diagonal elements
to be non-zero, the core tensor can represent both within- and
between-community interactions.

The elements of the core tensor are gamma-distributed,
\begin{align}
  \lambda_{c\rotcircle^k}^{(r)} &\sim
  \Gamma\!\left(\eta_c^{\rotcircle} \eta_c^{\leftrightarrow}
  \nu_k \rho_r, \delta \right) &\\
  \lambda_{\cdk}^{(r)} &\sim
  \Gamma\!\left(\eta_c^{\leftrightarrow} \eta_d^{\leftrightarrow}
  \nu_k \rho_r,
  \delta \right) &c \neq d.
\end{align}
Each community $c \in [C]$ has two positive weights
$\eta_c^{\rotcircle}$ and $\eta_c^{\leftrightarrow}$ that capture its
rates of within- and between-community interaction, respectively. Each
topic $k \in [K]$ has a positive weight $\nu_k$, while each regime $r
\in [R]$ has a positive weight $\rho_r$. We place an uninformative
prior over the within-community interaction rates and gamma shrinkage
priors over the other weights: $\eta_c^{\rotcircle} \sim
\Gamma\!\left(\epsilon_0, \epsilon_0\right)$,
$\eta_c^{\leftrightarrow} \sim \Gamma\!\left( \gamma_0 \,/\, C,
\zeta\right)$, $\nu_k \sim \Gamma\!\left( \gamma_0 \,/\, K,
\zeta\right)$, and $\rho_r \sim \Gamma\!\left( \gamma_0 \,/\, R,
\zeta\right)$. These priors bias BPTD toward learning latent structure
that is sparse. Finally, we assume that $\delta$ and $\zeta$ are drawn
from an uninformative gamma prior: $\delta, \zeta \sim
\Gamma\!\left(\epsilon_0, \epsilon_0\right)$.

As $K \rightarrow \infty$, the topic weights and their corresponding
action--topic factors constitute a draw $G_K = \sum_{k=1}^{\infty}
\nu_k\, \mathds{1}_{\boldsymbol{\phi}_k}$ from a gamma
process~\citep{ferguson73bayesian}. Similarly, as $R \rightarrow
\infty$, the regime weights and their corresponding time-step--regime
factors constitute a draw $G_R = \sum_{r=1}^{\infty} \rho_r\,
\mathds{1}_{\boldsymbol{\psi}_r}$ from another gamma process. As $C
\rightarrow \infty$, the within- and between-community interaction
weights and their corresponding country--community factors constitute
a draw $G_C = \sum_{c=1}^{\infty}
\eta_c^{\leftrightarrow}\,\mathds{1}_{\boldsymbol{\theta}_c}$ from a
marked gamma process~\citep{kingman72poisson}. The mark associated
with atom $\boldsymbol{\theta}_c = \left(\theta_{1c}, \ldots,
\theta_{V\!c}\right)$ is $\eta_c^{\rotcircle}$. We can view the
elements of the core tensor and their corresponding factors as a draw
$G = \sum_{c=1}^{\infty} \sum_{d=1}^{\infty} \sum_{k=1}^{\infty}
\sum_{r=1}^{\infty} \lambda_{\cdk}^{(r)} \,
\mathds{1}_{\boldsymbol{\theta}_c,\boldsymbol{\theta}_d,
  \boldsymbol{\phi}_k, \boldsymbol{\psi}_r}$ from a gamma process,
provided that the expected sum of the core tensor elements is finite.
This multirelational gamma process extends the relational gamma
process of \citet{zhou15infinite}.

\textbf{Proposition 1:} \emph{In the limit as $C, K, R \rightarrow
  \infty$, the expected sum of the core tensor elements is finite and
  equal to}
\begin{equation*}
  \mathbb{E}\left[\sum_{c=1}^{\infty}\sum_{k=1}^{\infty}\sum_{r=1}^{\infty}\!
    \left( \lambda_{c\rotcircle^k}^{(r)} +\! \sum_{d \neq c}
    \lambda_{\cdk}^{(r)}\right) \right] = \frac{1}{\delta} {\left(
    \frac{\gamma^3_0}{\zeta^3} + \frac{\gamma^4_0}{\zeta^4}\right)\!}.
\end{equation*}
We prove this proposition in the supplementary material.

\section{Connections to Previous Work}
\label{sec:connections}

\textbf{Poisson CP decomposition:} \citet{dubois10modeling} developed
a model that assigns each event token (ignoring time steps) to one of
$Q$ latent classes, where each class $q \in [Q]$ is characterized by
three categorical distributions---$\btheta^{\rightarrow}_q$ over
senders, $\btheta^{\leftarrow}_q$ over receivers, and $\bphi_q$ over
actions---i.e.,
\begin{equation}
  \label{eqn:mpmm}
  P(e_n \teq (\ija, t) \g z_n \teq q) =
  \theta^{\rightarrow}_{iq}\,\theta^{\leftarrow}_{jq}\,\phi_{aq}.
\end{equation}
This model is closely related to the Poisson-based model
of~\citet{schein15bayesian}, which explicitly uses the canonical
polyadic (CP) tensor decomposition~\citep{harshman70foundations} to
factorize count tensor $\boldsymbol{Y}$ into four latent factor
matrices. These factor matrices jointly embed senders, receivers,
action types, and time steps into a $Q$-dimensional space,
\begin{equation}
  \yijat \sim \Pois\! \left( \sum_{q=1}^Q
  \theta^{\rightarrow}_{iq}\,\theta^{\leftarrow}_{jq}\,\phi_{aq}\,\psi_{tq}
  \right),
\end{equation}
where $\theta^{\rightarrow}_{iq}$, $\theta^{\leftarrow}_{jq}$,
$\phi_{aq}$, and $\psi_{tq}$ are positive real numbers.

\citeauthor{schein15bayesian}'s model generalizes Bayesian Poisson
matrix
factorization~\citep{cemgil09bayesian,gopalan14bayesian,gopalan15scalable,zhou15negative}
and non-Bayesian Poisson CP
decomposition~\citep{chi12tensors,welling01positive}.

Although \citeauthor{schein15bayesian}'s model is expressed in terms
of a tensor of event type counts, the relationship between the
multinomial and Poisson distributions~\citep{kingman72poisson} means
that we can also express it in terms of a set of event tokens. This
yields an equation that is similar to equation~\ref{eqn:mpmm},
\begin{equation}
  P(e_n \teq (\ija, t) \g z_n \teq q) \propto
  \theta^{\rightarrow}_{iq}\,\theta^{\leftarrow}_{jq}\,\phi_{aq}\,\psi_{tq}.
\end{equation}
Conversely, \citeauthor{dubois10modeling}'s model can be expressed as
a CP tensor decomposition. This equivalence is analogous to the
relationship between Poisson matrix
factorization 
and latent Dirichlet allocation~\citep{blei03latent}.

We can make \citeauthor{schein15bayesian}'s model nonparametric by
adding a per-class positive weight $\lambda_q \sim
\Gamma(\frac{\gamma_0}{Q}, \zeta)$, i.e.,
\begin{equation}
  \yijat \sim \Pois\! \left( \sum_{q=1}^Q
  \theta^{\rightarrow}_{iq}\,\theta^{\leftarrow}_{jq}\,\phi_{aq}\,\psi_{tq}\,\lambda_q
  \right).
\end{equation}
As $Q \rightarrow \infty$ the per-class weights and their
corresponding latent factors constitute a draw from a gamma
process.

Adding this per-class weight reveals that CP decomposition is a
special case of Tucker decomposition where the cardinalities of the
latent dimensions are equal and the off-diagonal elements of the core
tensor are zero. \citeauthor{dubois10modeling}'s and
\citeauthor{schein15bayesian}'s models are therefore highly
constrained special cases of BPTD that cannot capture
dimension-specific structure, such as communities of countries or
topics of action types. These models require each latent class to
jointly summarize information about senders, receivers, action types,
and time steps. This requirement conflates communities of countries
and topics of action types, thus forcing each class to capture
potentially redundant information. Moreover, by definition, CP
decomposition models cannot express between-community interactions and
cannot express sender--receiver asymmetry without learning completely
separate latent factor matrices for senders and receivers. These
limitations make it hard to interpret these models as learning
community memberships.

\textbf{Infinite relational models:} The infinite relational model
(IRM) of \citet{kemp06learning} also learns latent structure specific
to each dimension of an $M$-dimensional tensor; however, unlike BPTD,
the elements of this tensor are binary, indicating the presence or
absence of the corresponding event type. The IRM therefore uses a
Bernoulli likelihood. \citet{schmidt13nonparametric} extended the IRM
to model a tensor of event counts by replacing the Bernoulli
likelihood with a Poisson likelihood (and gamma priors):
\begin{equation}
  \yijat \sim \Pois\! \left( \lambda^{(z_t)}_{z_i {\xrightarrow{z_a}}
    z_j} \right),
  \label{eqn:gpirm_likelihood}
\end{equation}
where $z_i, z_j \in [C]$ are the respective community assignments of
countries $i$ and $j$, $z_a \in [K]$ is the topic assignment of action
$a$, and $z_t \in [R]$ is the regime assignment of time step $t$. This
model, which we refer to as the gamma--Poisson IRM (GPIRM), allocates
$M$-dimensional event types to $M$-dimensional latent classes---e.g.,
it allocates all tokens of type $(\ija, t)$ to class $(z_i
{\xrightarrow{z_a}} z_j, z_t)$.

The GPIRM is a special case of BPTD where the rows of the latent
factor matrices are constrained to be ``one-hot'' binary
vectors---i.e., $\theta_{ic} \teq \mathds{1}(z_i \teq c)$,
$\theta_{jd} \teq \mathds{1}(z_j \teq d)$, $\phi_{ak} \teq
\mathds{1}(z_a \teq k)$, and $\psi_{tr} \teq \mathds{1}(z_t \teq
r)$.  With this constraint, the Poisson rates in
equations~\ref{eqn:likelihood} and~\ref{eqn:gpirm_likelihood} are
equal. Unlike BPTD, the GPIRM is a single-membership model. In
addition, it cannot express heterogeneity in rates of activity of the
countries, action types, and time steps. The latter limitation can be
remedied by letting $\theta_{i z_i}$, $\theta_{j z_j}$, $\phi_{a
  z_a}$, and $\psi_{t z_t}$ be positive real numbers. We refer to this
variant of the GPIRM as the degree-corrected GPIRM (DCGPIRM).

\textbf{Stochastic block models:} The IRM itself generalizes the
stochastic block model (SBM) of~\citet{nowicki01estimation}, which
learns latent structure from binary networks. Although the SBM was
originally specified using a Bernoulli likelihood,
\citet{karrer11stochastic} introduced an alternative specification
that uses the Poisson likelihood:
\begin{equation}
  \label{eqn:poisson_sbm_likelihood}
  y_{i {\xrightarrow{}} j} \sim \Pois\!\left(\sum_{c=1}^C
  \theta_{ic}\!  \sum_{d=1}^C \theta_{jd} \, \lambda_{c
    {\xrightarrow{}} d}\right),
\end{equation}
where $\theta_{ic} \teq \mathds{1}(z_i \teq c)$, $\theta_j \teq
\mathds{1}(z_j \teq d)$, and $\lambda_{c {\xrightarrow{}} d}$ is a
positive real number. Like the IRM and the GPIRM, the SBM is a
single-membership model and cannot express heterogeneity in the
countries' rates of activity. \citet{airoldi08mixed} addressed the
former limitation by letting $\theta_{ic} \in [0,1]$ such that
$\sum_{c=1}^C \theta_{ic} = 1$. Meanwhile, \citet{karrer11stochastic}
addressed the latter limitation by allowing both $\theta_{i z_i}$ and
$\theta_{j z_j}$ to be positive real numbers, much like the
DCGPIRM. \citet{ball11efficient} simultaneously addressed both
limitations by letting $\theta_{ic}, \theta_{jd} \geq 0$, but
constrained $\lambda_{c {\xrightarrow{}} d} \teq \lambda_{d
  {\xrightarrow{}} c}$. Finally, \citet{zhou15infinite} extended
\citeauthor{ball11efficient}'s model to be nonparametric and
introduced the Poisson--Bernoulli distribution to link binary data to
the Poisson likelihood in a principled fashion. In this model, the
elements of the core matrix and their corresponding factors constitute
a draw from a relational gamma process.

\textbf{Non-Poisson Tucker decomposition:} Researchers sometimes
refer to the Poisson rate in equation~\ref{eqn:poisson_sbm_likelihood}
as being ``bilinear'' because it can equivalently be written as
$\boldsymbol{\theta}_{j}\, \boldsymbol{\Lambda}\,
\boldsymbol{\theta}_{i}^{\top}$. \citet{nickel12factorizing}
introduced RESCAL---a non-probabilistic bilinear model for binary data
that achieves state-of-the-art performance at relation
extraction. \citet{nickel15review} then introduced several extensions for
extracting relations of different types. Bilinear models, such as
RESCAL and its extensions, are all special cases (albeit
non-probabilistic ones) of Tucker decomposition.

Hoff (\citeyear{hoff15equivariant}) recently developed a
Gaussian-based Tucker decomposition model and multilinear tensor
regression model~\citep{hoff14multilinear} for analyzing interaction
event data.

Finally, there are many other Tucker decomposition
methods~\citep{kolda09tensor}. Although these include
nonparametric~\citep{xu12infinite} and nonnegative
variants~\citep{kim07nonnegative,morup08algorithms,cichocki09nonnegative},
BPTD is the first such model to use a Poisson likelihood.

\section{Posterior Inference}
\label{sec:inference}

Given an observed count tensor $\boldsymbol{Y}$, inference in BPTD
involves ``inverting'' the generative process to obtain the posterior
distribution over the parameters conditioned on $\boldsymbol{Y}$ and
hyperparameters $\epsilon_0$ and $\gamma_0$. The posterior
distribution is analytically intractable; however, we can approximate
it using a set of posterior samples. We draw these samples using Gibbs
sampling, repeatedly resampling the value of each parameter from its
conditional posterior given $\boldsymbol{Y}$, $\epsilon_0$,
$\gamma_0$, and the current values of the other parameters. We express
each parameter's conditional posterior in a closed form using
gamma--Poisson conjugacy and the auxiliary variable techniques of
\citet{zhou12augment-and-conquer}. We provide the conditional
posteriors in the supplementary material.

The conditional posteriors depend on $\boldsymbol{Y}$ via a set of
``latent sources''~\citep{cemgil09bayesian} or subcounts. Because of
the Poisson additivity theorem~\cite{kingman72poisson}, each latent
source $y^{(tr)}_{ic {\xrightarrow{ak}} jd}$ is a Poisson-distributed
random variable:
\begin{align}
  \label{eqn:latent_source}
  y^{(tr)}_{ic {\xrightarrow{ak}} jd} &\sim \Pois\!\left( \theta_{ic}
  \, \theta_{jd} \,
  \phi_{ak}\,\psi_{tr}\,\lambda^{(r)}_{\cdk}\right)\\
  \label{eqn:latent_source_sum}
  y^{(t)}_{\ija} &= \sum_{c=1}^C \sum_{d=1}^D \sum_{k=1}^K
  \sum_{r=1}^R y^{(tr)}_{ic {\xrightarrow{ak}} jd}.
\end{align}
Together, equations~\ref{eqn:latent_source}
and~\ref{eqn:latent_source_sum} are equivalent to
equation~\ref{eqn:likelihood}. In practice, we can equivalently view
each latent source in terms of the token representation described in
section~\ref{sec:model},
\vspace{-0.4em}
\begin{equation}
  y^{(tr)}_{ic {\xrightarrow{ak}} jd} = \sum_{n=1}^N \mathds{1}(e_n
  \teq (\ija, t))\mathds{1}(z_n \teq (\cdk, r)),
  \label{eqn:latent_source_allocation}
\end{equation}
where each token's class assignment $z_n$ is an auxiliary latent
variable. Using this representation, computing the latent sources
(given the current values of the model parameters) simply involves
allocating event tokens to classes, much like the inference algorithm
for \citeauthor{dubois10modeling}'s model, and aggregating them using
equation~\ref{eqn:latent_source_allocation}. The conditional posterior
for each token's class assignment is
\begin{align}
&P(z_n \teq (\cdk, r) \g e_n \teq (\ija, t), \boldsymbol{Y}, \epsilon_0,
\gamma_0, \ldots) \notag \\
\label{eqn:z_n_gibbs_update}
&\qquad \propto \theta_{ic}\,\theta_{jd}\,\phi_{ak}\,\psi_{tr}\,\lambda^{(r)}_{\cdk}.
\end{align}
Computation is dominated by the normalizing constant
\begin{equation}
  \label{eqn:non-compositional}
  Z^{(t)}_{\ija} = \sum_{c=1}^C \sum_{d=1}^C \sum_{k=1}^K \sum_{r=1}^R
  \theta_{ic}\,\theta_{jd}\,\phi_{ak}\,\psi_{tr}\,\lambda^{(r)}_{\cdk}.
\end{equation}
Computing this normalizing constant na\"{i}vely involves $O(C \times C
\times K \times R)$ operations; however, because each latent class
$(\cdk, r)$ is composed of four separate dimensions, we can improve
efficiency. We instead compute
\begin{equation}
  \label{eqn:compositional}
  Z^{(t)}_{\ija} = \sum_{c=1}^C \theta_{ic} \!\sum_{d=1}^C
  \theta_{jd}\!\sum_{k=1}^K \theta_{ak}\!\sum_{r=1}^R \psi_{tr}
  \,\lambda^{(r)}_{\cdk},
\end{equation}
which involves $O(C + C + K + R)$ operations.

Compositional allocation using equations~\ref{eqn:z_n_gibbs_update}
and~\ref{eqn:compositional} improves computational efficiency
significantly over na\"{i}ve non-compositional allocation using
equations~\ref{eqn:z_n_gibbs_update} and \ref{eqn:non-compositional}.
In practice, we set $C$, $K$, and $R$ to large values to approximate
the nonparametric interpretation of BPTD. If, for example, $C \teq
50$, $K \teq 10$, and $R \teq 5$, computing the normalizing constant
for equation~\ref{eqn:z_n_gibbs_update} using
equation~\ref{eqn:non-compositional} requires 2,753 times the number
of operations implied by equation~\ref{eqn:compositional}.

\textbf{Proposition 2:} \emph{For an $M$-dimensional core tensor with
  $D_1 \times \ldots \times D_M$ elements, computing the normalizing
  constant using non-compositional allocation requires $1 \leq \pi <
  \infty$ times the number of operations required to compute it using
  compositional allocation. When $D_1 \teq \ldots \teq D_M \teq 1$,
  $\pi \teq 1$. As $D_m, D_{m'} \rightarrow \infty$ for any $m$ and
  $m'\!\neq\! m$, $\pi \rightarrow \infty$.}

We prove this proposition in the supplementary material.

\begin{figure}[t]
\centering
\includegraphics[width=0.90\columnwidth]{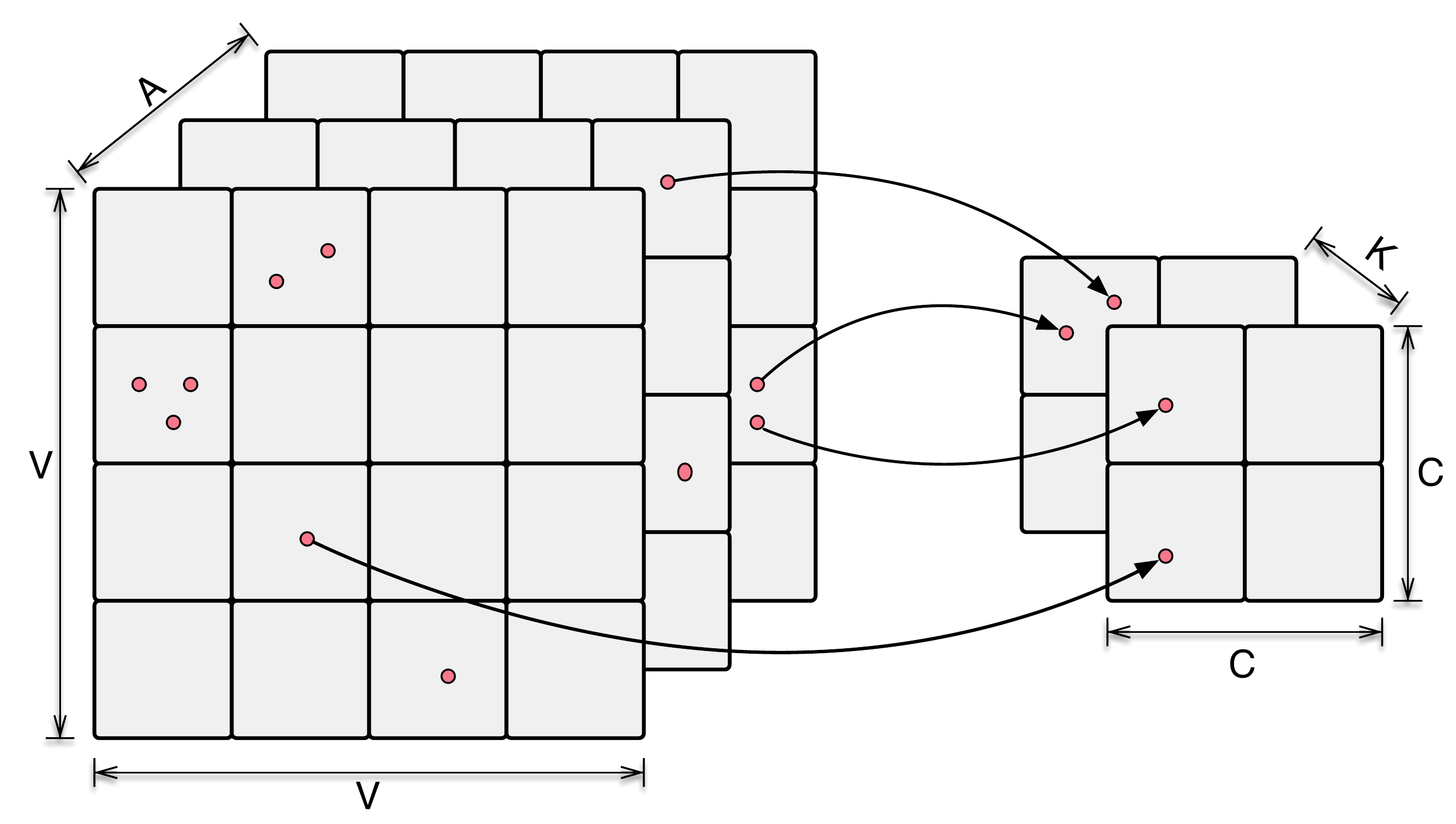}
\caption{Compositional allocation. For clarity, we show the allocation
  process for a three-dimensional count tensor (ignoring time
  steps). Observed three-dimensional event tokens (left) are
  compositionally allocated to three-dimensional latent classes
  (right).}
\label{fig:compositional}
\vskip -5pt
\end{figure}


BPTD and other Poisson-based models yield allocation inference
algorithms that take advantage of the inherent sparsity of the data
and scale with the number of event tokens. In contrast, non-Poisson
tensor decomposition models (including
\citeauthor{hoff15equivariant}'s model) lead to algorithms that scale
with the size of the count tensor. Allocation-based inference in BPTD
is especially efficient because it \emph{compositionally} allocates
each $M$-dimensional event token to an $M$-dimensional latent
class. Figure~\ref{fig:compositional} illustrates this process. CP
decomposition models, such as those of \citet{dubois10modeling} and
\citet{schein15bayesian}, only permit non-compositional
allocation. For example, while BPTD allocates each token $e_n \teq
(\ija, t)$ to a four-dimensional latent class $(\cdk, r)$,
\citeauthor{schein15bayesian}'s model allocates $e_n$ to a
one-dimensional latent class $q$ that cannot be decomposed. Therefore,
when $Q \teq C \times C \times K \times R$, BPTD yields a faster
allocation inference algorithm than \citeauthor{schein15bayesian}'s
model.

\section{Country--Country Interaction Event Data}
\label{sec:data}

Our data come from the Integrated Crisis Early Warning System (ICEWS)
of \citet{boschee??icews} and the Global Database of Events, Language,
and Tone (GDELT) of \citet{leetaru13gdelt}. ICEWS and GDELT both use
the Conflict and Mediation Event Observations (CAMEO)
hierarchy~\citep{gerner??conflict} for senders, receivers, and
actions.

The top-level CAMEO coding for senders and receivers is their country
affiliation, while lower levels in the hierarchy incorporate more
specific attributes like their sectors (e.g., government or civilian)
and their religious or ethnic affiliations. When studying
international relations using CAMEO-coded event data, researchers
usually consider only the senders' and receivers' countries. There are
249 countries represented in ICEWS, which include non-universally
recognized states, such as \emph{Occupied Palestinian Territory}, and
former states, such as \emph{Former Yugoslav Republic of Macedonia};
there are 233 countries in GDELT.

The top level for actions, which we use in our analyses, consists of
twenty action classes, roughly ranked according to their overall
sentiment. For example, the most negative is \emph{20---Use
  Unconventional Mass Violence}. CAMEO further divides these actions
into the QuadClass scheme: Verbal Cooperation (actions 2--5), Material
Cooperation (actions 6--7), Verbal Conflict (actions 8--16), and
Material Conflict (16--20). The first action (\emph{1---Make
  Statement}) is neutral.

\section{Predictive Analysis}
\label{sec:prediction}

\textbf{Baseline models:} We compared BPTD's predictive performance to
that of three baseline models, described in
section~\ref{sec:connections}: 1) GPIRM, 2) DCGPIRM, and 3) the
Bayesian Poisson tensor factorization (BPTF) model
of~\citet{schein15bayesian}. All three models use a Poisson likelihood
and have the same two hyperparameters as BPTD---i.e., $\epsilon_0$ and
$\gamma_0$. We set $\epsilon_0$ to 0.1, as recommended by
\citet{gelman06prior}, and we set $\gamma_0$ so that $\left(
\gamma_0\,/\, C\right)^2 \left( \gamma_0 \,/\, K \right)\left(
\gamma_0 \,/\, R\right) = 0.01$. This parameterization encourages the
elements of the core tensor $\boldsymbol{\Lambda}$ to be sparse. We
implemented an MCMC inference algorithm for each model. We provide the
full generative process for all three models in the supplementary
material.

GPIRM and DCGPIRM are both Tucker decomposition models and thus
allocate events to four-dimensional latent classes. The cardinalities
of these latent dimensions are the same as BPTD's---i.e., $C$, $K$,
and $R$. In contrast, BPTF is a CP decomposition model and thus
allocates events to one-dimensional latent classes. We set the
cardinality of this dimension so that the total number of latent
factors in BPTF's likelihood was equal to the total number of latent
factors in BPTD's likelihood---i.e., $Q \teq \lceil\frac{(V \times C)
  + (A \times K) + (T \times R) + (C^2 \times K \times R)}{V + V + A +
  T + 1}\rceil$. We chose not to let BPTF and BPTD use the same number
of latent classes---i.e., to set $Q \teq C^2 \times K \times R$. BPTF
does not permit non-compositional allocation, so MCMC inference
becomes very slow for even moderate values of $C$, $K$, and $R$. CP
decomposition models also tend to overfit when $Q$ is
large~\citep{zhao15bayesian}. Throughout our predictive experiments,
we let $C \teq 20$, $K \teq 6$, and $R \teq 3$.  These values were
well-supported by the data, as we explain in
section~\ref{sec:exploration}.

\textbf{Experimental setup:} We constructed twelve different observed
tensors---six from ICEWS and six from GDELT. Five of the six tensors
for each source (ICEWS or GDELT) correspond to one-year time spans
with monthly time steps, starting with 2004 and ending with 2008; the
sixth corresponds to a five-year time span with monthly time steps,
spanning 1995--2000. We divided each tensor $\boldsymbol{Y}$ into a
training tensor $\boldsymbol{Y}_{\textrm{train}} \teq
\boldsymbol{Y}^{(1)}, \ldots, \boldsymbol{Y}^{(T-3)}$ and a test
tensor $\boldsymbol{Y}_{\textrm{test}} \teq \boldsymbol{Y}^{(T-2)},
\ldots, \boldsymbol{Y}^{(T)}$. We further divided each test tensor
into a held-out portion and an observed portion via a binary mask. We
experimented with two different masks: one that treats the elements
involving the most active fifteen countries as the held-out portion
and the remaining elements as the observed portion, and one that does
the opposite. The first mask enabled us to evaluate the models'
reconstructions of the densest (and arguably most interesting) portion
of each test tensor, while the second mask enabled us to evaluate
their reconstructions of its complement. Across the entire GDELT
database, for example, the elements involving the most active fifteen
countries---i.e., 6\% of all 233 countries---account for 30\% of the
event tokens. Moreover, 40\% of these elements are non-zero. These
non-zero elements are highly dispersed, with a variance-to-mean ratio
of 220. In contrast, only 0.7\% of the elements involving the other
countries are non-zero. These elements have a variance-to-mean ratio
of 26.

For each combination of the four models, twelve tensors, and two
masks, we ran 5,000 iterations of MCMC inference on the training
tensor. We clamped the country--community factors, the action--topic
factors, and the core tensor and then inferred the time-step--regime
factors for the test tensor using its observed portion by running
1,000 iterations of MCMC inference. We saved every tenth sample after
the first 500. We used each sample, along with the country--community
factors, the action--topic factors, and the core tensor, to compute
the Poisson rate for each element in the held-out portion of the test
tensor. Finally, we averaged these rates across samples and used each
element's average rate to compute its probability. We combined the
held-out elements' probabilities by taking their geometric mean or,
equivalently, by computing their inverse perplexity. We chose this
combination strategy to ensure that the models were penalized heavily
for making poor predictions on the non-zero elements and were not
rewarded excessively for making good predictions on the zero
elements. By clamping the country--community factors, the
action--topic factors, and the core tensor after training, our
experimental setup is analogous to that used to assess collaborative
filtering models' strong generalization
ability~\citep{marlin04collaborative}.

\begin{figure*}[t]
\vspace{0.1em}
\centering
\includegraphics[width=\linewidth]{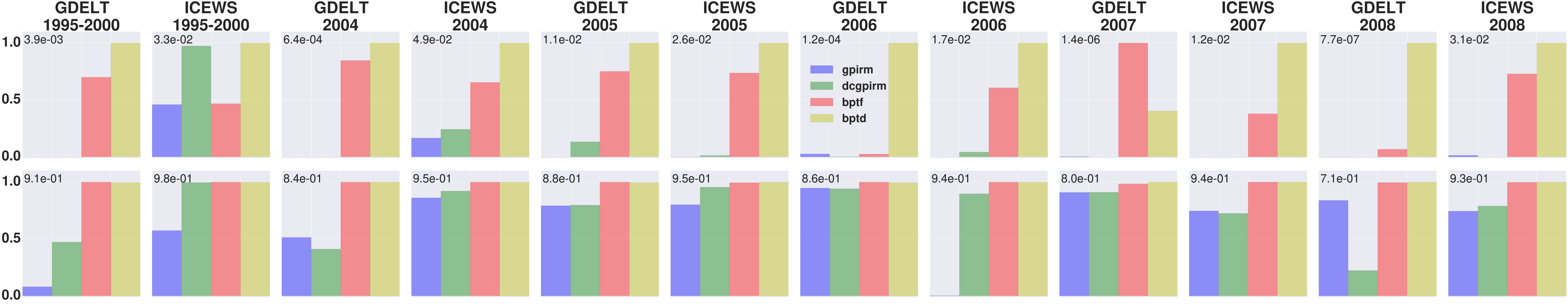}
\caption{Predictive performance. Each plot shows the inverse
  perplexity (higher is better) for the four models: the GPIRM (blue),
  the DCGPIRM (green), BPTF (red), and BPTD (yellow). In the
  experiments depicted in the top row, we treated the elements
  involving the most active countries as the held-out portion; in the
  experiments depicted in the bottom row, we treated the remaining
  elements as the held-out portion. For ease of comparison, we scaled
  the inverse perplexities to lie between zero and one; we give the
  scales in the top-left corners of the plots. BPTD outperformed the
  baselines significantly when predicting the denser portion of each
  test tensor (top row).}
\label{fig:prediction}
\vskip -5pt
\end{figure*}

\textbf{Results:} Figure~\ref{fig:prediction} illustrates the results
for each combination of the four models, twelve tensors, and two
masks. The top row contains the results from the twelve experiments
involving the first mask, where the elements involving the most active
fifteen countries were treated as the held-out portion. BPTD
outperformed the baselines significantly. BPTF---itself a
state-of-the-art model---performed better than BPTD in only one
experiment. In general, the Tucker decomposition allows BPTD to learn
richer latent structure that generalizes better to held-out data. The
bottom row contains the results from the experiments involving the
second mask. The models' performance was closer in these experiments,
probably because of the large proportion of easy-to-predict zero
elements. BPTD and BPTF performed indistinguishably in these
experiments, and both models outperformed the GPIRM and the
DCGPIRM. The single-membership nature of the GPIRM and the DCGPIRM
prevents them from expressing high levels of heterogeneity in the
countries' rates of activity. When the held-out elements were highly
dispersed, these models sometimes made extremely inaccurate
predictions. In contrast, the mixed-membership nature of BPTD and BPTF
allows them to better express heterogeneous rates of activity.

\section{Exploratory Analysis}
\label{sec:exploration}

We used a tensor of ICEWS events spanning 1995--2000, with monthly
time steps, to explore the latent structure discovered by BPTD. We
initially let $C \teq 50$, $K \teq 8$, and $R \teq 3$---i.e., $C
\times C \times K \times R = 60,000$ latent classes---and used the
shrinkage priors to adaptively learn the most appropriate numbers of
communities, topics, and regimes. We found $C \teq 20$ communities and
$K \teq 6$ topics with weights that were significantly greater than
zero. We provide a plot of the community weights in the supplementary
material. Although all three regimes had non-zero weights, one had a
much larger weight than the other two. For comparison,
\citet{schein15bayesian} used fifty latent classes to model the same
data, while \citet{hoff15equivariant} used $C \teq 4$, $K \teq 4$, and
$R \teq 4$ to model a similar tensor from GDELT.

\begin{figure*}[ht]
\vspace{-1.5em}
\centering
\begin{subfigure}[t]{0.24\linewidth}
\caption*{}
\includegraphics[width=\linewidth]{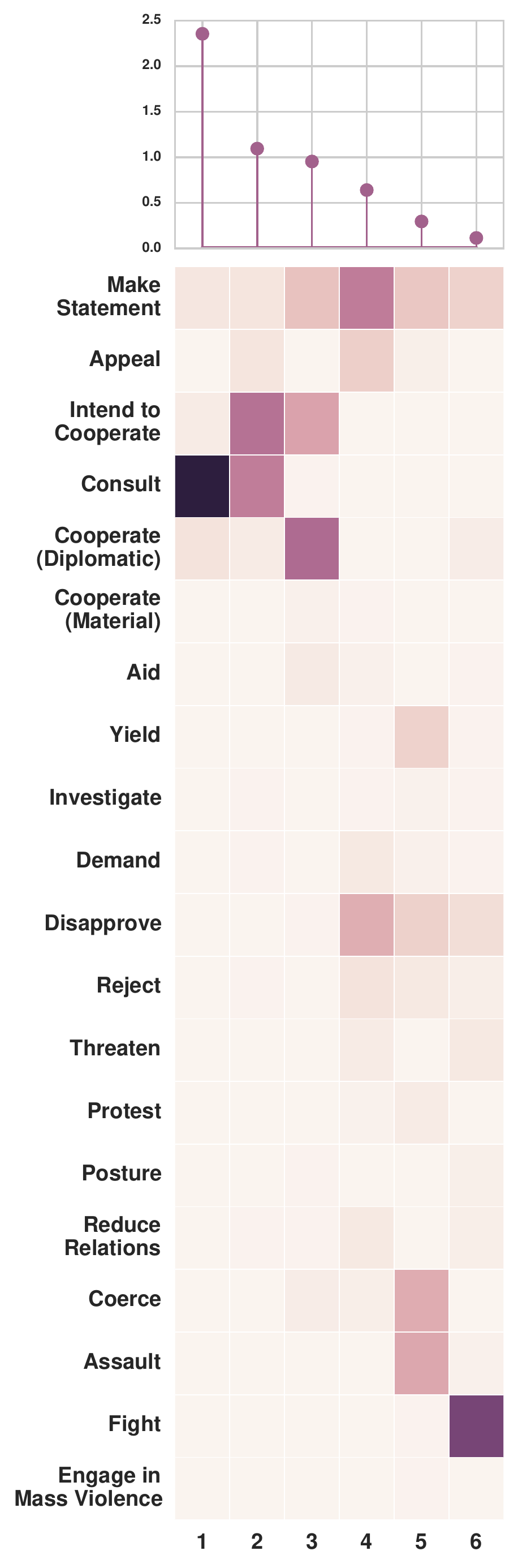}
\end{subfigure}
\begin{subfigure}[t]{0.74\linewidth}
\caption*{}
\includegraphics[width=\linewidth]{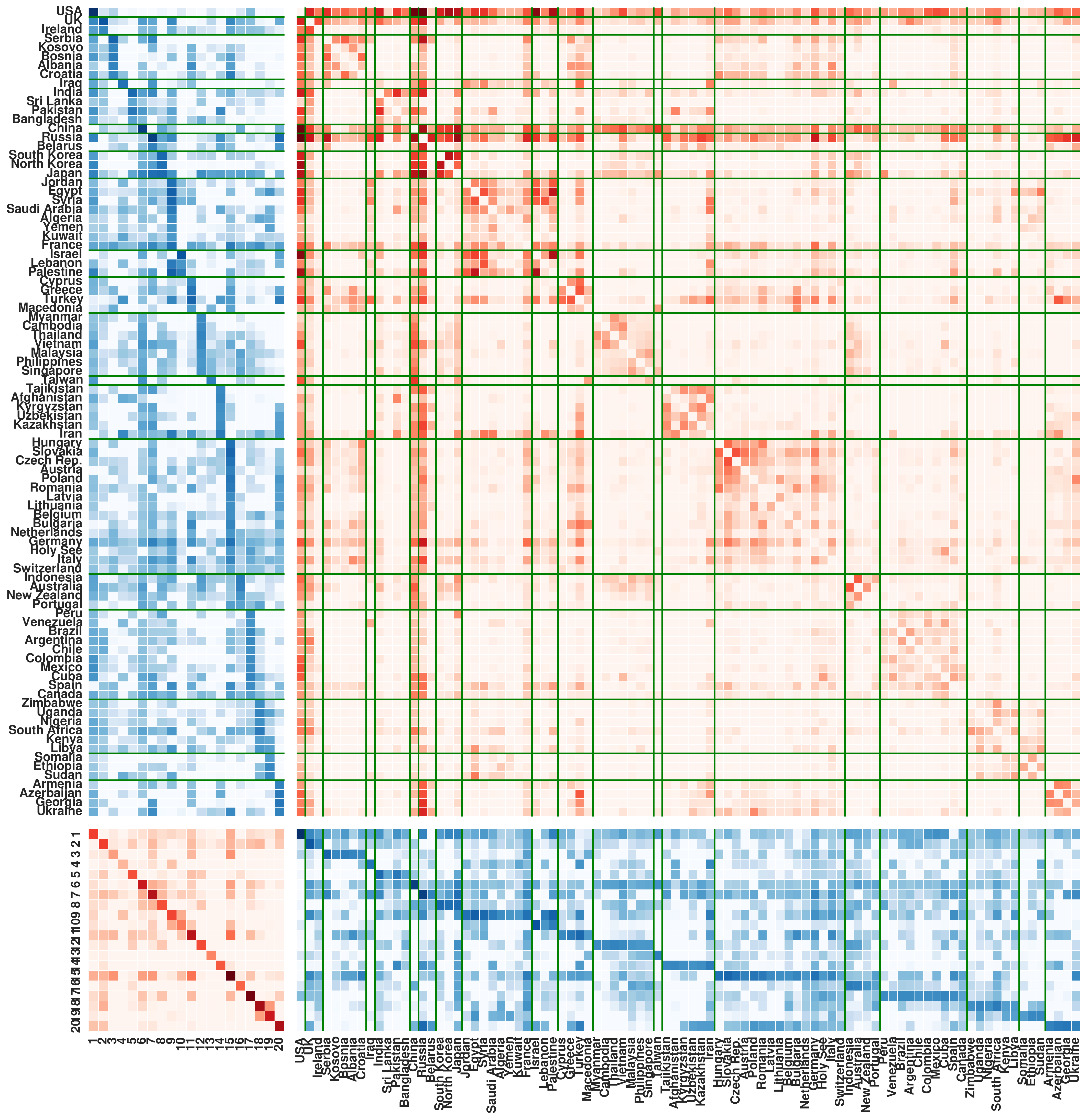}
\end{subfigure}
\caption{\emph{Left:} Action--topic factors. The topics are ordered by
  $\nu_1, \ldots, \nu_K$ (above the heatmap).  \emph{Right:} Latent
  structure discovered by BPTD for topic $k \teq 1$ and the most
  active regime, including the community--community interaction
  network (bottom left), the rate at which each country acts as a
  sender (top left) and a receiver (bottom right) in each community,
  and the number of times each country $i$ took an action associated
  with topic $k$ toward each country $j$ during regime $r$ (top
  right). We show only the most active 100 countries.}
\label{fig:exploration}
\vskip -5pt
\end{figure*}

\textbf{Topics of action types:} We show the inferred action--topic
factors as a heatmap in the left subplot of
figure~\ref{fig:exploration}. We ordered the topics by their weights
$\nu_1, \ldots, \nu_K$, which are above the heatmap. The inferred
topics correspond very closely to CAMEO's QuadClass scheme. Moving
from left to right, the topics place their mass on increasingly
negative actions. Topics 1 and 2 place most of their mass on Verbal
Cooperation actions; topic 3 places most of its mass on Material
Cooperation actions and the neutral \emph{1---Make Statement} action;
topic 4 places most of its mass on Verbal Conflict actions and the
\emph{1---Make Statement} action; and topics 5 and 6 place their mass
on Material Conflict actions.

\textbf{Topic-partitioned community--community networks:} In the right
subplot of figure~\ref{fig:exploration}, we visualize the inferred
community structure for topic $k \teq 1$ and the most active regime
$r$. The bottom-left heatmap is the community--community interaction
network $\boldsymbol{\Lambda}^{(r)}_k$. The top-left heatmap depicts
the rate at which each country $i$ acts as a sender in each community
$c$---i.e., $\theta_{ic} \sum_{j=1}^V \sum_{d=1}^C \theta_{jd}\,
\lambda^{(r)}_{\cdk}$. Similarly, the bottom-right heatmap depicts the
rate at which each country acts as a receiver in each community. The
top-right heatmap depicts the number of times each country $i$ took an
action associated with topic $k$ toward each country $j$ during regime
$r$---i.e., $\sum_{c=1}^C \sum_{d=1}^C \sum_{a=1}^A \sum_{t=1}^T
y^{(tr)}_{ic {\xrightarrow{ak}} jd}$. We grouped the countries by
their strongest community memberships and ordered the communities by
their within-community interaction weights $\eta_1^{\rotcircle},
\ldots, \eta_C^{\rotcircle}$, from smallest to largest; the thin green
lines separate the countries that are strongly associated with one
community from the countries that are strongly associated with its
adjacent communities.

Some communities contain only one or two strongly associated
countries. For example, community 1 contains only the US, community 6
contains only China, and community 7 contains only Russia and
Belarus. These communities mostly engage in between-community
interaction. Other larger communities, such as communities 9 and 15,
mostly engage in within-community interaction. Most communities have a
strong geographic interpretation. Moving upward from the bottom, there
are communities that correspond to Eastern Europe, East Africa,
South-Central Africa, Latin America, Australasia, Central Europe,
Central Asia, etc. The community--community interaction network
summarizes the patterns in the top-right heatmap. This topic is
dominated by the \emph{4--Consult} action, so the network is
symmetric; the more negative topics have asymmetric
community--community interaction networks. We therefore hypothesize
that cooperation is an inherently reciprocal type of interaction. We
provide visualizations for the other five topics in the supplementary
material.

\section{Summary}
\label{sec:summary}

We presented Bayesian Poisson Tucker decomposition (BPTD) for learning
the latent structure of international relations from country--country
interaction events of the form ``country $i$ took action $a$ toward
country $j$ at time $t$.'' Unlike previous models, BPTD takes
advantage of all three representations of an interaction event data
set: 1) a set of event tokens, 2) a tensor of event type counts, and
3) a series of weighted multinetwork snapshots. BPTD uses a Poisson
likelihood, respecting the discrete nature of the data and its
inherent sparsity. Moreover, BPTD yields a compositional allocation
inference algorithm that is more efficient than non-compositional
allocation algorithms. Because BPTD is a Tucker decomposition model,
it shares parameters across latent classes. In contrast, CP
decomposition models force each latent class to capture potentially
redundant information. BPTD therefore ``does more with less.'' This
efficiency is reflected in our predictive analysis: BPTD outperforms
BPTF---a CP decomposition model---as well as two other baselines. BPTD
learns interpretable latent structure that aligns with well-known
concepts from the networks literature. Specifically, BPTD learns
latent country--community memberships, including the number of
communities, as well as directed community--community interaction
networks that are specific to topics of action types and temporal
regimes. This structure captures the complexity of country--country
interactions, while revealing patterns that agree with our knowledge
of international relations. Finally, although we presented BPTD in the
context of interaction events, BPTD is well suited to learning latent
structure from other types of multidimensional count data.

\section*{Acknowledgements}

We thank Abigail Jacobs and Brandon Stewart for helpful
discussions. This work was supported by NSF \#SBE-0965436,
\#IIS-1247664, \#IIS-1320219; ONR \#N00014-11-1-0651; DARPA
\#FA8750-14-2-0009, \#N66001-15-C-4032; Adobe; the John Templeton
Foundation; the Sloan Foundation; the UMass Amherst Center for
Intelligent Information Retrieval. Any opinions, findings,
conclusions, or recommendations expressed in this material are the
authors' and do not necessarily reflect those of the sponsors.

\bibliography{references}

\begin{thebibliography}{36}
\providecommand{\natexlab}[1]{#1}
\providecommand{\url}[1]{\texttt{#1}}
\expandafter\ifx\csname urlstyle\endcsname\relax
  \providecommand{\doi}[1]{doi: #1}\else
  \providecommand{\doi}{doi: \begingroup \urlstyle{rm}\Url}\fi

\bibitem[Airoldi et~al.(2008)Airoldi, Blei, Feinberg, and Xing]{airoldi08mixed}
Airoldi, E.~M., Blei, D.~M., Feinberg, S.~E., and Xing, E.~P.
\newblock Mixed membership stochastic blockmodels.
\newblock \emph{Journal of Machine Learning Research}, 9:\penalty0 1981--2014,
  2008.

\bibitem[Ball et~al.(2011)Ball, Karrer, and Newman]{ball11efficient}
Ball, B., Karrer, B., and Newman, M.~E.~J.
\newblock Efficient and principled method for detecting communities in
  networks.
\newblock \emph{Physical Review E}, 84\penalty0 (3), 2011.

\bibitem[Blei et~al.(2003)Blei, Ng, and Jordan]{blei03latent}
Blei, D., Ng, A., and Jordan, M.
\newblock Latent {D}irichlet allocation.
\newblock \emph{Journal of Machine Learning Research}, 3:\penalty0 993--1022,
  2003.

\bibitem[Boschee et~al.()Boschee, Lautenschlager, {O'B}rien, Shellman, Starz,
  and Ward]{boschee??icews}
Boschee, E., Lautenschlager, J., {O'B}rien, S., Shellman, S., Starz, J., and
  Ward, M.
\newblock {ICEWS} coded event data.
\newblock Harvard Dataverse.
\newblock V10.

\bibitem[Cemgil(2009)]{cemgil09bayesian}
Cemgil, A.~T.
\newblock Bayesian inference for nonnegative matrix factorisation models.
\newblock \emph{Computational Intelligence and Neuroscience}, 2009.

\bibitem[Chi \& Kolda(2012)Chi and Kolda]{chi12tensors}
Chi, E.~C. and Kolda, T.~G.
\newblock On tensors, sparsity, and nonnegative factorizations.
\newblock \emph{SIAM Journal on Matrix Analysis and Applications}, 33\penalty0
  (4):\penalty0 1272--1299, 2012.

\bibitem[Cichocki et~al.(2009)Cichocki, Zdunek, Phan, and
  i~Amari]{cichocki09nonnegative}
Cichocki, A., Zdunek, R., Phan, A.~H., and i~Amari, S.
\newblock \emph{Nonnegative Matrix and Tensor Factorizations: Applications to
  Exploratory Multi-Way Data Analysis and Blind Source Separation}.
\newblock John Wiley \& Sons, 2009.

\bibitem[Du{B}ois \& Smyth(2010)Du{B}ois and Smyth]{dubois10modeling}
Du{B}ois, C. and Smyth, P.
\newblock Modeling relational events via latent classes.
\newblock In \emph{Proceedings of the Sixteenth ACM SIGKDD International
  Conference on Knowledge Discovery and Data Mining}, pp.\  803--812, 2010.

\bibitem[Ferguson(1973)]{ferguson73bayesian}
Ferguson, T.~S.
\newblock A {B}ayesian analysis of some nonparametric problems.
\newblock \emph{The Annals of Statistics}, 1\penalty0 (2):\penalty0 209--230,
  1973.

\bibitem[Gelman(2006)]{gelman06prior}
Gelman, A.
\newblock Prior distributions for variance parameters in hierarchical models.
\newblock \emph{Bayesian Analysis}, 1\penalty0 (3):\penalty0 515--533, 2006.

\bibitem[Gerner et~al.()Gerner, Schrodt, Abu-Jabr, and
  Yilmaz]{gerner??conflict}
Gerner, D.~J., Schrodt, P.~A., Abu-Jabr, R., and Yilmaz, {\"{O}}.
\newblock Conflict and mediation event observations ({CAMEO}): A new event data
  framework for the analysis of foreign policy interactions.
\newblock Working paper.

\bibitem[Gopalan et~al.(2014)Gopalan, Ruiz, Ranganath, and
  Blei]{gopalan14bayesian}
Gopalan, P., Ruiz, F.~J.~R., Ranganath, R., and Blei, D.~M.
\newblock Bayesian nonparametric {P}oisson factorization for recommendation
  systems.
\newblock In \emph{Proceedings of the Seventeenth International Conference on
  Artificial Intelligence and Statistics}, volume~33, pp.\  275--283, 2014.

\bibitem[Gopalan et~al.(2015)Gopalan, Hofman, and Blei]{gopalan15scalable}
Gopalan, P., Hofman, J., and Blei, D.
\newblock Scalable recommendation with {P}oisson factorization.
\newblock In \emph{Proceedings of the Thirty-First Conference on Uncertainty in
  Artificial Intelligence}, 2015.

\bibitem[Harshman(1970)]{harshman70foundations}
Harshman, R.
\newblock Foundations of the {PARAFAC} procedure: Models and conditions for an
  ``explanatory'' multimodal factor analysis.
\newblock \emph{UCLA Working Papers in Phonetics}, 16:\penalty0 1--84, 1970.

\bibitem[Hoff(2014)]{hoff14multilinear}
Hoff, P.
\newblock Multilinear tensor regression for longitudinal relational data.
\newblock arXiv:1412.0048, 2014.

\bibitem[Hoff(2015)]{hoff15equivariant}
Hoff, P.
\newblock Equivariant and scale-free {T}ucker decomposition models.
\newblock \emph{Bayesian Analysis}, 2015.

\bibitem[Karrer \& Newman(2011)Karrer and Newman]{karrer11stochastic}
Karrer, B. and Newman, M.~E.~J.
\newblock Stochastic blockmodels and community structure in networks.
\newblock \emph{Physical Review E}, 83\penalty0 (1), 2011.

\bibitem[Kemp et~al.(2006)Kemp, Tenenbaum, Griffiths, Yamada, and
  Ueda]{kemp06learning}
Kemp, C., Tenenbaum, J.~B., Griffiths, T.~L., Yamada, T., and Ueda, N.
\newblock Learning systems of concepts with an infinite relational model.
\newblock In \emph{Proceedings of the Twenty-First National Conference on
  Artificial Intelligence}, 2006.

\bibitem[Kim \& Choi(20007)Kim and Choi]{kim07nonnegative}
Kim, Y.-D. and Choi, S.
\newblock Nonnegative {T}ucker decomposition.
\newblock In \emph{Proceedings of the IEEE Conference on Computer Vision and
  Pattern Recognition}, 20007.

\bibitem[Kingman(1972)]{kingman72poisson}
Kingman, J.~F.~C.
\newblock \emph{Poisson Processes}.
\newblock Oxford University Press, 1972.

\bibitem[Kolda \& Bader(2009)Kolda and Bader]{kolda09tensor}
Kolda, T.~G. and Bader, B.~W.
\newblock Tensor decompositions and applications.
\newblock \emph{SIAM Review}, 51\penalty0 (3):\penalty0 455--500, 2009.

\bibitem[Leetaru \& Schrodt(2013)Leetaru and Schrodt]{leetaru13gdelt}
Leetaru, K. and Schrodt, P.
\newblock {GDELT}: Global data on events, location, and tone, 1979--2012.
\newblock Working paper, 2013.

\bibitem[Marlin(2004)]{marlin04collaborative}
Marlin, B.
\newblock Collaborative filtering: A machine learning perspective.
\newblock Master's thesis, University of Toronto, 2004.

\bibitem[M{\o}rup et~al.(2008)M{\o}rup, Hansen, and Arnfred]{morup08algorithms}
M{\o}rup, M., Hansen, L.~K., and Arnfred, S.~M.
\newblock Algorithms for sparse nonnegative {T}ucker decompositions.
\newblock \emph{Neural Computation}, 20\penalty0 (8):\penalty0 2112--2131,
  2008.

\bibitem[Nickel et~al.(2012)Nickel, Tresp, and Kriegel]{nickel12factorizing}
Nickel, M., Tresp, V., and Kriegel, H.-P.
\newblock Factorizing {YAGO}: Scalable machine learning for linked data.
\newblock In \emph{Proceedings of the Twenty-First International World Wide Web
  Conference}, pp.\  271--280, 2012.

\bibitem[Nickel et~al.(2015)Nickel, Murphy, Tresp, and
  Gabrilovich]{nickel15review}
Nickel, M., Murphy, K., Tresp, V., and Gabrilovich, E.
\newblock A review of relational machine learning for knowledge graphs: From
  multi-relational link prediction to automated knowledge graph construction.
\newblock arXiv:1503.00759, 2015.

\bibitem[Nowicki \& Snijders(2001)Nowicki and Snijders]{nowicki01estimation}
Nowicki, K. and Snijders, T.~A.~B.
\newblock Estimation and prediction for stochastic blockstructures.
\newblock \emph{Journal of the American Statistical Association}, 96\penalty0
  (455):\penalty0 1077--1087, 2001.

\bibitem[Schein et~al.(2015)Schein, Paisley, Blei, and
  Wallach]{schein15bayesian}
Schein, A., Paisley, J., Blei, D.~M., and Wallach, H.
\newblock {B}ayesian {Poisson} tensor factorization for inferrring multilateral
  relations from sparse dyadic event counts.
\newblock In \emph{Proceedings of the Twenty-First ACM SIGKDD International
  Conference on Knowledge Discovery and Data Mining}, pp.\  1045--1054, 2015.

\bibitem[Schmidt \& M{\o}rup(2013)Schmidt and M{\o}rup]{schmidt13nonparametric}
Schmidt, M.~N. and M{\o}rup, M.
\newblock Nonparametric {B}ayesian modeling of complex networks: An
  introduction.
\newblock \emph{IEEE Signal Processing Magazine}, 30\penalty0 (3):\penalty0
  110--128, 2013.

\bibitem[Tucker(1964)]{tucker64extension}
Tucker, L.~R.
\newblock The extension of factor analysis to three-dimensional matrices.
\newblock In Frederiksen, N. and Gulliksen, H. (eds.), \emph{Contributions to
  Mathematical Psychology}. Holt, Rinehart and Winston, 1964.

\bibitem[Welling \& Weber(2001)Welling and Weber]{welling01positive}
Welling, M. and Weber, M.
\newblock Positive tensor factorization.
\newblock \emph{Pattern Recognition Letters}, 22\penalty0 (12):\penalty0
  1255--1261, 2001.

\bibitem[Xu et~al.(2012)Xu, Yan, and Qi]{xu12infinite}
Xu, Z., Yan, F., and Qi, Y.
\newblock Infinite {T}ucker decomposition: Nonparametric {B}ayesian models for
  multiway data analysis.
\newblock In \emph{Proceedings of the Twenty-Ninth International Conference on
  Machine Learning}, pp.\  1023--1030, 2012.

\bibitem[Zhao et~al.(2015)Zhao, Zhang, and Cichocki]{zhao15bayesian}
Zhao, Q., Zhang, L., and Cichocki, A.
\newblock Bayesian {CP} factorization of incomplete tensors with automatic rank
  determination.
\newblock \emph{IEEE Transactions on Pattern Analysis and Machine
  Intelligence}, 37\penalty0 (9):\penalty0 1751--1763, 2015.

\bibitem[Zhou(2015)]{zhou15infinite}
Zhou, M.
\newblock Infinite edge partition models for overlapping community detection
  and link prediction.
\newblock In \emph{Proceedings of the Eighteenth International Conference on
  Artificial Intelligence and Statistics}, pp.\  1135--1143, 2015.

\bibitem[Zhou \& Carin(2012)Zhou and Carin]{zhou12augment-and-conquer}
Zhou, M. and Carin, L.
\newblock Augment-and-conquer negative binomial processes.
\newblock In \emph{Advances in Neural Information Processing Systems
  Twenty-Five}, pp.\  2546--2554, 2012.

\bibitem[Zhou \& Carin(2015)Zhou and Carin]{zhou15negative}
Zhou, M. and Carin, L.
\newblock Negative binomial process count and mixture modeling.
\newblock \emph{IEEE Transactions on Pattern Analysis and Machine
  Intelligence}, 37\penalty0 (2):\penalty0 307--320, 2015.

\end{thebibliography}
\bibliographystyle{icml2016}

\end{document}